\documentclass[conference]{IEEEtran}
\IEEEoverridecommandlockouts
\usepackage{cite}
\usepackage{amsmath,amssymb,amsfonts}
\usepackage{graphicx}
\usepackage{textcomp}
\usepackage{xcolor}

\usepackage{newfloat}
\usepackage{amsmath}
\usepackage{amsfonts}
\usepackage{graphicx}
\usepackage{amsmath}
\usepackage{amsfonts}
\usepackage{multirow}
\usepackage{tabularx}
\usepackage{array}
\usepackage{algorithm}
\usepackage{algpseudocode}
\usepackage{xspace}
\usepackage{subcaption} 
\usepackage[labelformat=empty]{subcaption}

\def\eg{\emph{e.g.,}\xspace}

\def\ie{\emph{i.e.,}\xspace}

\def\name{\textsc{\textit{Horus}}\xspace}
\newcommand{\one}{({\em i})\xspace}
\newcommand{\two}{({\em ii})\xspace}
\newcommand{\three}{({\em iii})\xspace}

\def\BibTeX{{\rm B\kern-.05em{\sc i\kern-.025em b}\kern-.08em
    T\kern-.1667em\lower.7ex\hbox{E}\kern-.125emX}}
\begin{document}

\title{Heterogeneity-Oblivious Robust Federated Learning\\
}


\author{
\IEEEauthorblockN{
Weiyao Zhang\textsuperscript{\dag},
Jinyang Li\textsuperscript{\dag},
Qi Song\textsuperscript{\dag\ddag},
Miao Wang\textsuperscript{\ddag},
Chungang Lin\textsuperscript{\dag\ddag},
Haitong Luo\textsuperscript{\dag\ddag},\\
Xuying Meng\textsuperscript{\dag\S\ddag\textasteriskcentered},
Yujun Zhang\textsuperscript{\dag\ddag\textasteriskcentered}
}
\IEEEauthorblockA{
\textsuperscript{\dag}Institute of Computing Technology, Chinese Academy of Sciences, China.
\textsuperscript{\S}Purple Mountain Laboratories, China.\\\textsuperscript{\ddag}University of Chinese Academy of Sciences, China.
\textsuperscript{\textasteriskcentered}Corresponding authors.
}
}

\maketitle

\begin{abstract}
Federated Learning (FL) remains highly vulnerable to poisoning attacks, especially under real-world hyper-heterogeneity, where clients differ significantly in data distributions, communication capabilities, and model architectures. Such heterogeneity not only undermines the effectiveness of aggregation strategies but also makes attacks more difficult to detect. Furthermore, high-dimensional models expand the attack surface. To address these challenges, we propose \name, a heterogeneity-oblivious robust FL framework centered on low-rank adaptations (LoRAs). Rather than aggregating full model parameters, \name inserts LoRAs into empirically stable layers and aggregates only LoRAs to reduce the attack surface.
We uncover a key empirical observation that the input projection (LoRA‑A) is markedly more stable than the output projection (LoRA‑B)  under heterogeneity and poisoning. Leveraging this, we design a Heterogeneity-Oblivious Poisoning Score using the features from LoRA‑A to filter poisoned clients.  For the remaining benign clients, we propose projection-aware  aggregation mechanism to preserve collaborative signals while suppressing drifts, which reweights client updates by consistency with the global directions. Extensive experiments across diverse datasets, model architectures, and attacks demonstrate that \name consistently outperforms state-of-the-art baselines in both robustness and accuracy.
\end{abstract}

\section{Introduction}
\label{sec:introduction}

Federated Learning (FL) has gained significant traction as a privacy-preserving paradigm for distributed training, enabling clients to collaboratively learn a global model without sharing their raw data~\cite{li2020federated,DBLP:conf/infocom/GaoZG025}. However, the decentralized nature of FL inherently introduces serious security vulnerabilities, making it susceptible to poisoning attacks, in which attackers inject malicious data or local updates. Such attacks pose a particularly insidious threat, as they can stealthily degrade or manipulate the global model over time~\cite{DBLP:conf/wacv/XuZH25}. For example, perturbing a federated model deployed in vehicular systems could autonomously start the vehicle or execute an emergency brake, thereby endangering human lives and compromising property safety~\cite{shejwalkar2021manipulating}.

These security challenges are amplified in real-world FL due to the presence of hyper-heterogeneity across clients.
\one Data heterogeneity. Due to differences in user behavior, environments, and tasks, local data distributions are highly non-IID. Attackers can exploit this by mimicking edge-case distributions, making malicious updates appear as plausible but atypical client behaviors~\cite{wan2024federated}.
\two Communication heterogeneity\cite{chen2024fedhello}. Clients differ in compute power, bandwidth, and availability\cite{liu2024fml}, leading to differing constraints on update frequency and model size. When benign clients communicate infrequently, poisoned updates may dominate aggregation for extended periods without timely correction by the benign clients.
\three Model architecture heterogeneity\cite{li2021federated}. Clients retain autonomy over their local models and may adopt fundamentally different architectures. For instance, a high-performance client may use a CNN-based model for image classification, while a resource-constrained edge device (\eg embedded camera) may transform images into sequences and utilize an RNN model instead\cite{visin2015renet}. 
Such architectural differences result in inconsistent update dimensions and semantics, making direct comparison infeasible and hindering unified detection of poisoned updates. 
Even worse, with hyper-heterogeneity, malicious deviations can be easily disguised as benign variations. 
Moreover, the high dimensionality of model updates further exacerbates these challenges, which not only increases the difficulty of poisoning detection, but also expands the attack surface by allowing adversaries to inject subtle yet effective perturbations along more vulnerable directions, which is known as ``the curse of dimensionality''~\cite{chen2023robustfl,shejwalkar2021manipulating}.

Under these challenges, existing methods often suffer from performance degradation or even collapse. For instance, robust FL such as DnC~\cite{shejwalkar2021manipulating}, FLDetector~\cite{zhang2022fldetector}, and LASA~\cite{xu2025lasa} have shown partial effectiveness under data heterogeneity, but remain constrained to homogeneous model architectures and balanced communication environments.
Even recent efforts that begin to address model heterogeneity typically focus on minor architectural variations within the same family (\eg ResNet vs. MobileNet), without supporting fundamentally different structures (\eg CNNs vs. RNNs) or highly imbalanced communication conditions~\cite{alam2022fedrolex,DBLP:journals/network/YuL21,diao2020heterofl}, and there is no relevant work to address the robustness issues brought about by the hyper-heterogeneity in FL.
Thus, there is a pressing need for a robust federated learning capable of operating under hyper-heterogeneity conditions. Such a framework should be:
\one \textbf{Heterogeneity-oblivious}: remain immune to hyper-heterogeneity, while supporting stable cross-client aggregation, and use stable signals to effectively distinguish poisoning updates from benign, heterogeneity-induced drifts.
\two \textbf{Dimension-reduction}: constrain updates to a low-rank subspace to shrink the high-dimensional attack surface and distill essential learning signals. At the same time, reduce payload to keep benign clients synchronized and prevent attackers from gaining an early advantage due to stragglers~\cite{DBLP:conf/allerton/MitliagkasZHR16}.

In light of the above, we propose a \underline{H}eterogeneity-\underline{O}blivious \underline{R}ob\underline{US}t federated learning framework \name centered on low-rank adaptations (LoRAs)~\cite{hu2022lora}. Rather than aggregating full model parameters, we insert LoRA into the stable layers of each client’s model and aggregate only these adaptations.  
LoRA is a parameter-efficient fine-tuning mechanism that freezes the backbone and constrains updates to a low-rank subspace, mitigating the curse of dimensionality and reducing communication load. More importantly, unlike pruning~\cite{DBLP:conf/nips/IvkinRUBSA19}, quantization~\cite{DBLP:conf/nips/AlistarhG0TV17}, or distillation~\cite{DBLP:journals/corr/abs-1910-03581}, which typically require retraining or global structural alignment, LoRA acts as a plug-in module, allowing heterogeneous clients to retain their own architectures.
To further harden the robustness, we first conduct an empirical analysis of LoRA stability across layers and clients. We find that the input projection (LoRA-A) remains consistently stable across rounds and clients, only weakly affected by hyper-heterogeneity. By contrast, the output projection (LoRA-B) is far more volatile, showing larger energy shifts and directional drift under both benign heterogeneity and poisoning perturbations, which makes it prone to false positives if used for detection.
Based on our observation, we functionally decouple the A from the LoRA and leverage it as a stable anchor for poisoning detection. We then design a heterogeneity-oblivious poisoning score based on the characteristics of LoRA-A to identify poisoned clients.
For aggregation, projection-aware aggregation is designed to align the model's dimension and aggregate according to the updates that are directionally consistent with the global trend.

To sum up, our main contributions are as follows:

\begin{itemize}

\item We propose the first robust framework for hyper-heterogeneous federated learning, which leverages low-rank adaptations to mitigate security risks associated with the curse of dimensionality. By decoupling LoRA modules, the framework supports heterogeneity-oblivious poisoning detection and aggregation, making it well-suited for realistic FL scenarios involving hyper-heterogeneity.

\item We conduct an in-depth analysis of the stability differences between LoRA-A and LoRA-B under poisoning attacks. Based on this insight, we design a heterogeneity-oblivious detection mechanism and a projection-aware aggregation strategy, enabling robust aggregation in the presence of hyper-heterogeneity.

\item We introduce a novel heterogeneity-oblivious poisoning score that is invariant to the dimensionality or shape of the original LoRA matrices, enabling consistent detection performance across structurally diverse clients.

\item Extensive experiments across multiple datasets, attack types, and heterogeneous configurations demonstrate that our method consistently outperforms state-of-the-art baselines in both robustness and accuracy.

\end{itemize}
\section{Relatedwork}
\label{Relatedwork}

\subsection{Robust Federated Learning}

Federated learning is inherently susceptible to poisoning attacks due to its decentralized structure and limited visibility into client-side behavior. In particular, model poisoning attacks involve adversarial clients uploading carefully crafted gradients or parameter updates to compromise the integrity of the global model~\cite{fang2020local, bagdasaryan2020backdoor}, while data poisoning corrupts local training datasets to induce systematic errors during aggregation~\cite{steinhardt2017certified}.
To address these threats, a variety of robust aggregation strategies have been proposed to mitigate poisoning behavior in federated learning. Approaches such as Krum~\cite{blanchard2017machine}, Mrum~\cite{blanchard2017machine}, Trimmed Mean~\cite{yin2018byzantine}, and Median~\cite{yin2018byzantine} filter out anomalous updates based on statistical distance or ranking heuristics, assuming that poisoned updates deviate significantly from benign ones.
More recent defenses go beyond simple distance metrics. For instance, FLDetector~\cite{zhang2022fldetector} detects malicious clients via checking their model-updates consistency across rounds. Dnc~\cite{shejwalkar2021manipulating} employs spectral analysis to detect and filter outliers in poisoned data, which is called divide-and-conquer. LASA~\cite{DBLP:conf/wacv/XuZH25} leverages a layer-wise filter that adaptively selects benign layers using both magnitude and direction metrics across all clients for aggregation.

Hyper-heterogeneity in federated learning makes poisoning detection challenging: clients may differ substantially in data distributions, model architectures, and parameter dimensions. Distance- or clustering-based detectors operating in parameter space are therefore brittle under such variability and often trigger high false positives.

\subsection{Heterogeneous and Efficient Federated Learning}
High-dimensional model updates enlarge the adversarial attack surface (``curse of dimensionality''), making coordinate-wise outliers and structured perturbations harder to filter.
One line of work, therefore, reduces the effective dimension before aggregation.
For example, LASA~\cite{DBLP:conf/wacv/XuZH25} ranks coordinates by saliency for each layer and aggregates only a sparse subset, thereby suppressing poisoning directions. However, it may over-trim benign diversity under hyper-heterogeneity.
In parallel, there are many efficiency-oriented techniques that can be paired with a robust detector. 
{Fjord}~\cite{yu2023fjord} uses ordered dropout to extract nested submodels from a large network via runtime pruning, facilitating heterogeneous deployment without retraining and supporting fair, accurate training across devices.
{FedRolex}~\cite{alam2022fedrolex} employs a rolling sub-model extraction scheme that allows different model parts to be trained.
{FedALA}~\cite{zhang2023fedala} adaptively aggregates the global model and local model on the selected layer, and {FedHello}~\cite{chen2024fedhello} enables collaboration across foundation models using heterogeneous LoRA.
{FGGP}~\cite{wan2024federated} learns personalized projection bases and aggregates them via graph-guided alignment. However, none of them can support the heterogeneous architecture. {HeteroFL}~\cite{diao2020heterofl} and {MFL}~\cite{DBLP:journals/network/YuL21} are efforts enabling clients with different architectures. They allocate subnetworks tailored to each client's capability while maintaining a shared global model, focusing on communication efficiency, but their clients belong to the same model family (ResNets with different sizes) rather than completely different architectures (CNNs vs. RNNs).

While these methods effectively address heterogeneity and dimension-reduction to some degree, they generally do not provide explicit defenses against poisoning, often assume compatible models, and can sacrifice robustness when faced with heterogeneity.

\section{Motivation}

In federated learning with hyper-heterogeneity, model updates from different clients are often high-dimensional, noisy, and semantically mismatched. These challenges severely hinder both poisoning detection and robust aggregation. Furthermore, exchanging full model parameters enlarges the attack surface in the high-dimensional space, exacerbating vulnerability due to the curse of dimensionality. All those call for a heterogeneity-oblivious, low-rank robust FL that stably aggregates, separates poison from benign drift, and compresses updates to cut attack surface and payload.

To this end, we adopt LoRA as a plug-in inserted into selected layers on each client~\cite{hu2022lora}. By constraining updates to a compact low-rank subspace, LoRA preserves the dominant directions of model evolution while compressing communication. Because LoRA is a layer-local and architecture-agnostic plug-in, heterogeneous clients can keep their native backbones yet exchange updates in a shared low-rank format, which shrinks the attack surface, reduces payload, and helps benign clients stay synchronized under communication constraints.

\begin{figure}[t]
    \centering

    \begin{minipage}[b]{0.48\linewidth}
        \centering
        \includegraphics[width=\linewidth]{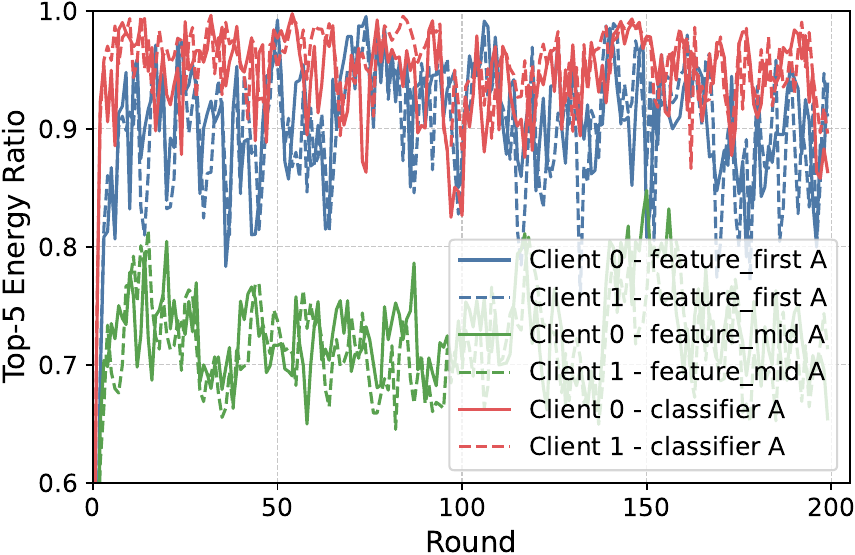}
        
    \end{minipage}
    \hfill
    \begin{minipage}[b]{0.48\linewidth}
        \centering
        \includegraphics[width=\linewidth]{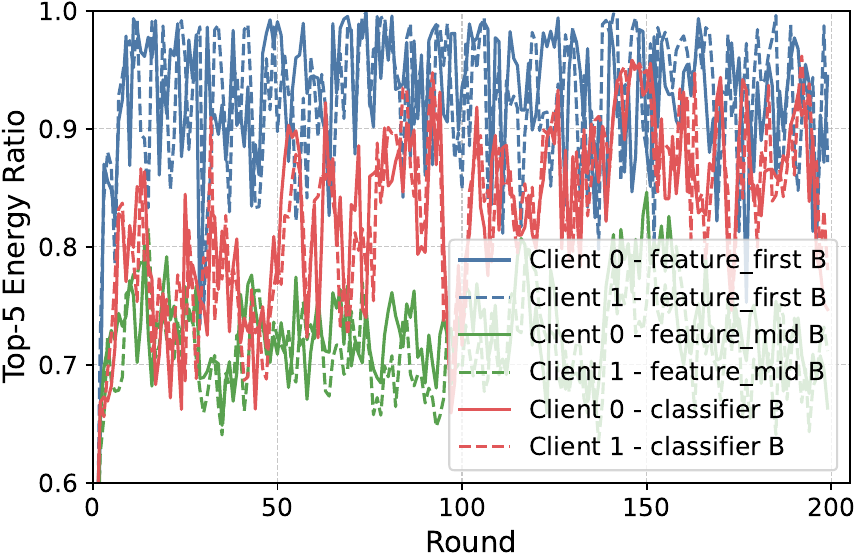}
        
    \end{minipage}

    \caption{Top-5 singular value energy ratio for LoRA-A (left) and LoRA-B (right) of \textit{feature\_first}, \textit{feature\_mid}, and \textit{classifier} from client 0 (poisoned) and client 1 (poisoned).}
    \label{fig:multi-layers_A_B_energy}
    \vspace{-10pt}
\end{figure}

However, as a plug-in, LoRA modifies a layer (base weight is $W$) by a low-rank increment without altering the layer's original input-output mapping, \eg $W'=W+\Delta W, \Delta W = B A$, $A \in \mathbb{R}^{r \times d_{\text{in}}}, B \in \mathbb{R}^{d_{\text{out}} \times r}$, where $A$ \& $B$ denote LoRA-A and LoRA-B, respectively, and $r \ll \min(d_{\text{in}}, d_{\text{out}})$, $d_{\text{in}}$ and $d_{\text{out}}$ are the input and output dimensions of the injected layer. As a result, LoRA inherits heterogeneity-induced inconsistencies. Naively averaging LoRA updates can remain unstable under hyper-heterogeneity. We therefore conduct extensive analyses to identify the stable components across clients, which in turn guide our heterogeneity-oblivious detection and robust aggregation design. Specifically, we use 10 clients spanning two architecture (CNNs and RNNs), and evaluate on CIFAR-10/100 and FMNIST (partitioned non-IID via different Dirichlet sampler), and analyze client behavior under benign and six representative poisoning attacks. More experimental details appear in \S~\ref{sec:evaluation}. Notably, figures in this section show one CIFAR10-LIE-$\alpha{=}0.5$ instance due to space, but the same qualitative trends hold across datasets, $\alpha$, architectures, and attacks. We summarize the key observations below.

\textit{{\textbf{Observation 1}:} The first-layer and the classifier layer exhibit more stability across heterogeneous clients, making them suitable candidates for robust federated learning.}

To evaluate the stability of LoRA under hyper-heterogeneity and poisoning attack, we functionally decouple LoRA as $A$ \& $B$, and perform singular value decomposition (SVD) on them. Then, \textit{Top-$k$ singular value energy ratio} is used to analyze the distribution of singular values in the $A$ and $B$ across different model layers and clients~\cite{eckart1936approximation}. This metric measures how much an update concentrates in $k$ dominant directions.
A high ratio means a small set of principal components carries most of the signal, making it easier to detect poisoning that break this directional pattern, whereas a low ratio indicates noisy updates that are more prone to heterogeneity or poisoning. The choice of $k$ follows the paper~\cite{eckart1936approximation}.

We visualize the Top-5 energy ratio of two poisoned clients (0 and 1) as an example across three layers: the first layer (\textit{feature-first} for encoding the input), a randomly selected middle layer (\textit{feature-mid}), and the \textit{classifier} layer. As shown in Fig.~\ref{fig:multi-layers_A_B_energy}, the \textit{feature-mid} exhibits a lower ratio, indicating poor cross-client stability, making it unsuitable for poisoning detection and robust aggregation. Such instability of \textit{feature-mid} may arise from that, it encodes high-level abstract features that are more sensitive to variations.
In contrast, \textit{feature-first} captures general low-level features and \textit{classifier} focuses on a compact task-specific projection. Both layers exhibit smoother temporal trends across clients, showing their robustness to heterogeneity and realizing the detectability of poisoning during aggregation. Motivated by these insights, we perform robust FL on the \textit{feature-first} and \textit{classifier}.




\textit{\textbf{Observation 2:} LoRA-A is relatively insensitive to poisoning perturbations yet provides a stable and discriminative signal for identifying poisoned updates, whereas LoRA-B is more volatile and prone to false positives.}


\begin{figure}[t]
    \centering

    \begin{minipage}[b]{0.48\linewidth}
        \centering
        \includegraphics[width=\linewidth]{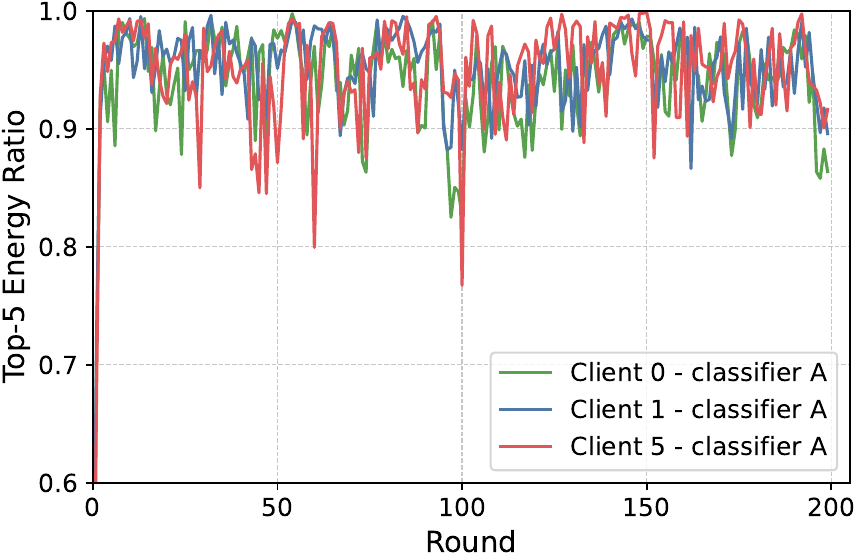}
        
    \end{minipage}
    \hfill
    \begin{minipage}[b]{0.48\linewidth}
        \centering
        \includegraphics[width=\linewidth]{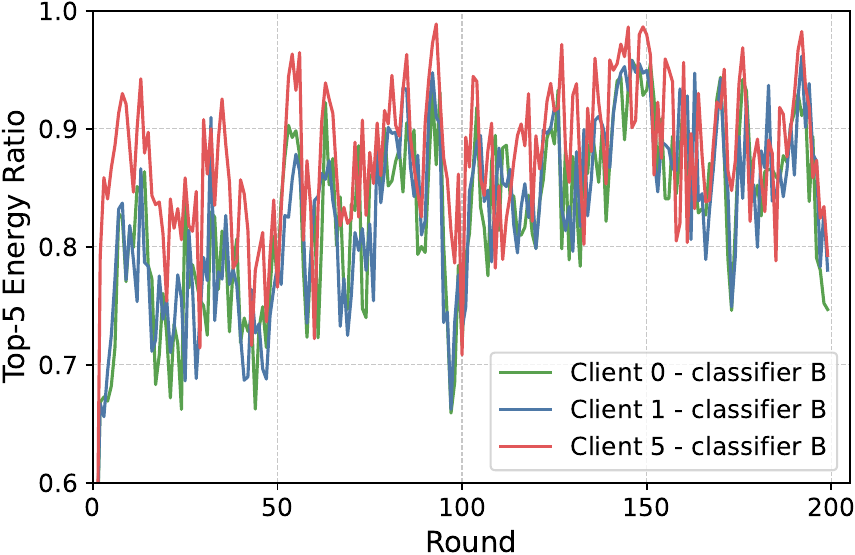}
        
    \end{minipage}

    \caption{Top-5 singular value energy ratio for LoRA-A (left) and LoRA-B (right) of \textit{classifier} from client 0 (poisoned), client 1 (benign), and client 5 (poisoned).}
    \label{fig:classifier_A_B_energy}
    \vspace{-10pt}
\end{figure}

\begin{figure*}
    \centering
    \includegraphics[width=0.82\linewidth]{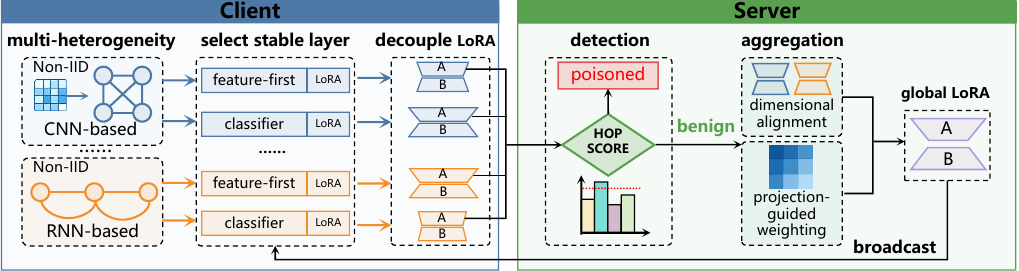}
    \caption{The framework of our proposed \name}
    \label{fig:framework}
    \vspace{-15pt}
\end{figure*}

To identify the stable components within LoRA, we apply poisoning attacks to two random clients (client 0 and client 5 for example), while keeping client 1 benign. We then track the evolution of the Top-5 energy ratio across rounds. Fig.~\ref{fig:classifier_A_B_energy} illustrates this trend using the \textit{classifier} as an example. After the attack is initiated, both LoRA-A and LoRA-B exhibit certain fluctuations. However, LoRA-A remains consistently stable. It is always stably maintained at over 90\%, indicating that the dominant update directions in A are largely preserved. 
Notably, the energy of poisoned clients (0 and 5) in LoRA-A is always lower than that of the benign client 1, suggesting that LoRA-A provides a distinguishable signal for detecting poisoning deviations.
In contrast, LoRA-B shows significantly higher volatility across both benign and poisoned clients. In some cases, 
the energy of benign or poisoned clients has no discrimination, which could lead to elevated false-positive rates if LoRA-B is used directly for detection.

Overall, in FL with hyper-heterogeneity, poisoning detection and robust aggregation require stability and a low false-positive rate rather than overly sensitive responses. The experiments in \S\ref{sec:ablation} also validate this insight. Therefore, \textit{feature-first} and \textit{classifier} can be used for aggregation under hyper-heterogeneity, and their LoRA-A can serve as a reliable signal for poisoning detection.

\section{Preliminary}





Federated learning enables distributed clients to collaboratively train without exposing raw data. To address the robustness challenges brought by the hyper-heterogeneity, according to our observation, we insert trainable low-rank adaptations into specific stable layers (\ie \textit{feature-first} and \textit{classifier}). In LoRA, given a original weight $W$, its update $\Delta W$ is confined to a structured subspace by $\Delta W = B A$, 
where $A$ and $B$ denote LoRA-A and LoRA-B, respectively. Each client retains a private backbone, and only LoRAs are shared during training, which is free from the communication restriction and shrinks the attack surface.

For $N$ participating clients, after local training, each client $c$ retains their private model backbone, and only shares $(A_c, B_c)$. The server uses a detection algorithm to kick out the poisoned clients, and applies an aggregation rule \( \mathcal{A}(\cdot) \) to compute:
\begin{equation}
\small
\bar{A} = \mathcal{A}(\{ A_c\}_{c=1}^{N}), \quad \bar{B} = \mathcal{A}(\{ B_c\}_{c=1}^{N})  
\end{equation}

The aggregated $\bar{A}, \bar{B}$ are then broadcast for local update. \S\ref{sec:methodology} details how to design the detection and aggregation rule. 

Empirically, we find that LoRA-A exhibits stronger stability, while LoRA-B is more sensitive to heterogeneity. Hence, we functionally decouple them: $A$ is used for poisoning detection, while $A$ \& $B$ support aggregation. This separation enables effective detection and aggregation under hyper-heterogeneity.

\section{Methodology}
\label{sec:methodology}

\subsection{Framework}
To enable robust FL under hyper-heterogeneity, we propose \name, a novel framework built on LoRA. In each round (Fig.~\ref{fig:framework}), participating clients update their models with private data. Instead of uploading full model parameters, each client extracts LoRA updates from two empirically stable layers (\ie \textit{feature-first} and \textit{classifier}). These updates are then decoupled into LoRA-A and LoRA-B components and transmitted to the server. Upon receiving the updates, the server performs a two-stage process (Algorithm~\ref{alg:HORUS}):

\one Heterogeneity-Oblivious Poisoning Detection: We exploit the cross‑client stability of LoRA‑A as the detection signal and define a Heterogeneity‑Oblivious Poisoning Score by fusing spectral features from its singular values. Clients with scores above an adaptive threshold are discarded.

\two Projection-Guided Aggregation: For benign clients, LoRAs are first aligned in dimension. The server then prioritizes LoRAs that align with global directions. This mitigates adversarial drift while preserving benign diversity.

The aggregated update is broadcast to participating clients and re-integrated into their local models for the next round.

\subsection{Heterogeneity-Oblivious Poisoning Detection}

\subsubsection{Heterogeneity-Oblivious Poisoning Score}
Hyper-heterogeneity in federated learning misleads poisoning detection. To address this, we propose the \textit{Heterogeneity-Oblivious Poisoning Score} (HOPS), a detection metric designed for hyper-heterogeneous FL. HOPS operates purely on the singular value of LoRA-A, making it agnostic to model architecture, layer shape, or semantics. It combines two spectral indicators:

\begin{itemize}
    \item \textit{Spectral Entropy}:
The spectral entropy $\mathcal{H}$ quantifies the global dispersion of energy across singular modes: a higher entropy suggests a flatter spectrum with no dominant direction, while a lower entropy indicates the existence of principal directions. When a poisoning perturbation is dispersive (\eg LIE~\cite{baruch2019little}), it tends to increase the spectral entropy, making the attack easier to detect. Given the singular values $\sigma_1, \dots, \sigma_r$ of a client's LoRA-A (rank $r$). We define the normalized energy distribution:
\begin{equation}
\small
\tilde{\sigma}_i \;=\; \frac{\sigma_i}{\sum_{j=1}^{r}\sigma_j},
\qquad
\mathcal{H} \;=\; -\sum_{i=1}^{r} \tilde{\sigma}_i \log \tilde{\sigma}_i.
\end{equation}

\item \textit{Top-$k$ Energy Ratio: }However, there also exist directional attacks (\eg Min-Sum~\cite{shejwalkar2021manipulating}), which deliberately concentrate their perturbations along the dominant directions of model updates to evade detection while still inducing model corruption. Therefore, assessing the energy concentration in the main directions is essential. To this end, we compute the top-$k$ energy ratio as a complementary metric, where $k \ll r$, and larger $\mathcal{R}_k$ suggests a focus on dominant directions:
\begin{equation}
\small
\mathcal{R}_k = \frac{\sum_{i=1}^k \sigma_i}{\sum_{j=1}^r \sigma_j}.
\end{equation}

\end{itemize}

To jointly account for dominant direction and the whole matrix's dispersion, we introduce a novel combination of two indicators, defined by their absolute deviations from round-wise reference statistics (line 10 in Algorithm~\ref{alg:HORUS}).
\begin{equation}
\small
\label{equa:HOPS}
S_c
\;=\;
\lambda\,\big| \big(1 - \mathcal{R}_k\big) - \mu_{\mathcal{R}} \big|
\;+\;
(1-\lambda)\,\left| \frac{\mathcal{H} - \mu_{\mathcal{H}}}{\sigma_{\mathcal{H}}} \right|,
\end{equation}
where $\mu_{\mathcal{R}}$ is the mean of $1-\mathcal{R}_k$ across clients in the current round, and $\mu_{\mathcal{H}},\sigma_{\mathcal{H}}$ are the round-wise mean and standard deviation of entropy. The weight $\lambda\in[0,1]$ balances the complementary microscopic concentration and macroscopic dispersion. 
HOPS is sensitive to deviations in either direction, whether directional attacks drive $\mathcal{R}_k\uparrow$ and $\mathcal{H}\downarrow$, or dispersive attacks cause $\mathcal{R}_k\downarrow$ and $\mathcal{H}\uparrow$, HOPS scores absolute deviations from the round's mean, so both extremes are flagged. Moreover, because HOPS depends only on singular values (architecture/shape-agnostic) and uses per-round normalization, it remains stable under hyper-heterogeneity.


\subsubsection{Detection Rule}
At round $t$, the server computes $S_{c,t}$ for each participating client $c$ and sets an adaptive threshold
\begin{equation}
\small
\theta_t \;=\; \mathrm{Percentile}_p\big(\{S_{c,t}\}_{c}\big)
\end{equation}
\eg, $p=95$ as in~\cite{shejwalkar2021manipulating}. Clients with $S_{c,t}>\theta_t$ are flagged and excluded from aggregation, ensuring that only reliable updates contribute to the global model (line 12 in Algorithm~\ref{alg:HORUS}).

\subsection{Projection-Guided Aggregation}

Under hyper-heterogeneity, the dimensions of selective layers differ across clients. Even with the rank $r$, LoRAs remain structurally heterogeneous because $d_{\text{in}}$ and $d_{\text{out}}$ depend on the inserted layer. To enable effective aggregation under such heterogeneity, we carry out two-phase work.

\subsubsection{Dimensional Alignment}

For client $c$, direct element-wise aggregation is infeasible the sizes of  $A_c\!\in\!\mathbb{R}^{r\times d_{\text{in}}^{(c)}}$ and $B_c\!\in\!\mathbb{R}^{d_{\text{out}}^{(c)}\times r}$ are layer-specific. Thus, we zero-pad each matrix to the global maximum shape: $\tilde{A}_c \in \mathbb{R}^{r \times d_{\text{in}}^{\max}}$, $\qquad 
\tilde{B}_c \in \mathbb{R}^{d_{\text{out}}^{\max} \times r}$,
where $d_{\text{in}}^{\max}$ and $d_{\text{out}}^{\max}$ are the maximum input/output dimensions across clients. However, the padded regions are meaningless, and direct averaging padded LoRA dilutes the meaningful updates. To prevent zero-padding from skewing the result, we introduce binary masks $M_c^A, M_c^B \in \{0,1\}^{\text{matching shape}}$, with $1$ on valid entries and $0$ on padded entries. Masked averaging is then performed element-wise:
\begin{equation}
\label{equa:alignment}
\small
\bar{A} = \frac{\sum_{c=1}^{N} \tilde{A}_c \odot M_c^A}{\sum_{c=1}^{N} M_c^A}, \quad
\bar{B} = \frac{\sum_{c=1}^{N} \tilde{B}_c \odot M_c^B}{\sum_{c=1}^{N} M_c^B}.
\end{equation}

\noindent where $\odot$ and the division are element-wise. Entries with zero denominators are ignored by the mask (\ie not aggregated).

\subsubsection{Projection-Guided Weighting} 
To mitigate poisoning drifts while preserving benign diversity, and further defend against attacks related to the principal direction, such as direction‑flip attack~\cite{shejwalkar2021manipulating}, we introduce a projection-based weighting mechanism and compute per-client directions and project them onto the global trend. First, we perform singular value decomposition (SVD) on the LoRA:
\begin{equation}
\small
\tilde{A}_c = U_c^A \Sigma_c^A (V_c^A)^\top,\qquad 
\tilde{B}_c = U_c^B \Sigma_c^B (V_c^B)^\top,
\end{equation}
and take the first right singular vectors $v_c^{A,(1)} := V_c^A[:,1]$ and $v_c^{B,(1)} := V_c^B[:,1]$, which capture the dominant update directions with maximal energy. Let $v_{g}^{A,(1)}$ and $v_{g}^{B,(1)}$ be the global directions from the previous round. We define consistency weights by the projection magnitudes (line 17 in Algorithm~\ref{alg:HORUS}):
\begin{equation}
\label{equa:projection_weight}
\small
\alpha_c^A \;=\; \big|\langle v_c^{A,(1)},\, v_{g}^{A,(1)} \rangle\big|,\qquad
\alpha_c^B \;=\; \big|\langle v_c^{B,(1)},\, v_{g}^{B,(1)} \rangle\big|,
\end{equation}
$\alpha_c^A,\alpha_c^B\in[0,1]$, and they are applied to masked averaging:
\begin{equation}
\small
\bar{A}
= \frac{\sum_{c=1}^{N} \alpha_c^A \,\big(\tilde{A}_c \odot M_c^A\big)}
        {\sum_{c=1}^{N} \alpha_c^A \, M_c^A},
\qquad
\bar{B}
= \frac{\sum_{c=1}^{N} \alpha_c^B \,\big(\tilde{B}_c \odot M_c^B\big)}
        {\sum_{c=1}^{N} \alpha_c^B \, M_c^B}.
\label{equa:weighted_aggregation}
\end{equation}

\begin{algorithm}[!t]
\small
\caption{Heterogeneity-Oblivious Robust Federated Learning}
\label{alg:HORUS}
\begin{algorithmic}[1]
\Require Clients $\mathcal{C}$, LoRA rank $r$, top-$k$, percentile $p$, rounds $T$, local epochs $E$; 
         initial global directions $v_g^{A,(1)}, v_g^{B,(1)}$
\Ensure Global LoRA $(\bar{A}, \bar{B})$
\Statex \textbf{[Client-side Poisoning Detection and Aggregation]}
\For{round $t=1$ to $T$}
  \For{each client $c \in \mathcal{C}$}
    \State Receive $(\bar{A}, \bar{B})$ from server, trim to local shapes.
    \State Local train for $E$ epochs on private data
    \State Decouple LoRA of stable layers as $A_c, B_c$
    \State Send $A_c, B_c$ to server
  \EndFor
  \Statex \textbf{[Server-side Poisoning Detection and Aggregation]}
\For{round $t=1$ to $T$}
\State Receive all $A, B$ from clients
    \State Compute HOPS of each $c \in \mathcal{C}$ by Eq.~(\ref{equa:HOPS})
  \State Set threshold $\theta_t$ as the $p$-th percentile of $\{S_c\}$
  \State Identify benign clients $\mathcal{C}_{\text{benign}} \!\!=\!\{\,c \mid S_c < \theta_t\,\}$
  \For{each $c \in \mathcal{C}_{\text{benign}}$}
      \State Dimensional alignment by Eq. (\ref{equa:alignment})
    \State Computing weight based on Eq. (\ref{equa:projection_weight})
  \EndFor
  \State Aggregate LoRA-A and LoRA-B by Eq. (\ref{equa:weighted_aggregation})
  \State Broadcast to all participating clients
\EndFor
\EndFor
\end{algorithmic}
\end{algorithm}

\subsubsection{Broadcast}
After Eq. (\ref{equa:weighted_aggregation}), the server broadcasts $\bar{A}$ and $\bar{B}$ to all participating clients. 
Under hyper-heterogeneity, projection-guided aggregation bridges architectural and semantic disparities and preserves directionally consistent collaboration while mitigating poisoning-induced drifts. Then, each client trims the global update to its local shape and updates its LoRA for the next round.


\section{Evaluation}
\label{sec:evaluation}

\begin{table*}[t]
\caption{Global accuracy of baselines and \name on three datasets with different $\alpha$ of Dirichlet distribution}
\small
\centering
\resizebox{\linewidth}{!}{
\renewcommand{\arraystretch}{1.15}
\begin{tabular}{c|>{\centering\arraybackslash}p{37pt}|llllllllllllllllll}
\hline
\multirow{2}{*}{AGR} & \multirow{2}{*}{Dataset}  & \multicolumn{3}{c}{label-flip}                                                   & \multicolumn{3}{c}{LIE}                                                         & \multicolumn{3}{c}{min-max}                                                      & \multicolumn{3}{c}{min-sum}                                                     & \multicolumn{3}{c}{Fang}                                                        & \multicolumn{3}{c}{AGR-tailored}                                                \\ \cline{3-20} 
                     &                           & \multicolumn{1}{c}{0.10} & \multicolumn{1}{c}{0.30} & \multicolumn{1}{c|}{0.50} & \multicolumn{1}{c}{0.10} & \multicolumn{1}{c}{0.30} & \multicolumn{1}{c|}{0.50} & \multicolumn{1}{c}{0.10}  & \multicolumn{1}{c}{0.30} & \multicolumn{1}{c|}{0.50} & \multicolumn{1}{c}{0.10} & \multicolumn{1}{c}{0.30} & \multicolumn{1}{c|}{0.50} & \multicolumn{1}{c}{0.10} & \multicolumn{1}{c}{0.30} & \multicolumn{1}{c|}{0.50} & \multicolumn{1}{c}{0.10}  & \multicolumn{1}{c}{0.30} & \multicolumn{1}{c}{0.50} \\ \hline
Krum                 & \multirow{9}{*}{CIFAR10}  & 30.72                    & 32.13                    & 39.53                     & 27.50                    & 31.72                    & 36.46                     & 27.77                     & 29.16                    & 32.88                     & 30.67                    & 33.38                    & 38.76                     & 29.19                    & 31.98                    & 36.75                     & 29.81                     & 31.26                    & 41.70                    \\
MKrum                &                           & 31.36                    & 37.50                    & 41.20                     & 31.62                    & 35.27                    & 40.12                     & 25.24                     & 29.00                    & 36.40                     & 27.41                    & 30.02                    & 38.27                     & 25.80                    & 33.55                    & 41.88                     & 27.41                     & \underline{34.62}                    & 39.15                    \\
Bulyan               &                           & 34.55                    & 38.18                    & 44.56                     & 32.13                    & 35.84                    & 46.67                     & 28.32                     & 30.70                    & 35.54                     & 31.25                    & 32.62                    & 37.23                     & 34.64                    & \underline{37.61}                    & 42.50                     & 34.41                     & 33.28                    & 40.59                    \\
Median               &                           & 37.74                    & 39.69                    & 47.62                     & 31.81                    & 37.92                    & 40.90                     & 29.02                     & \underline{36.83}                    & 45.25                     & 34.16                    & 31.94                    & 38.27                     & 29.96                    & 32.49                    & 41.10                     & 30.12                     & 32.31                    & 37.63                    \\
Trmean               &                           & 38.06                    & 43.85                    & 49.52                     & 24.69                    & 32.72                    & 40.06                     & 26.22                     & 28.27                    & 37.84                     & 31.35                    & 32.58                    & 40.57                     & 27.82                    & 31.70                    & 39.17                     & 28.67                     & 27.15                    & 34.78                    \\
Dnc                  &                           & 35.83                    & 44.00                    & \underline{54.70}                     & 28.69                    & 32.16                    & \underline{48.39}                     & 24.26                     & 35.38                    & \underline{46.31}                     & 27.70                    & 33.15                    & 43.16                     & 28.69                    & 34.03                    & 39.49                     & \multicolumn{1}{c}{28.32} & 32.71                    & 40.81                    \\
FLDetecter           &                           & 39.13                    & 43.93                    & 53.02                     & 30.63                    & 37.34                    & 40.24                     & 26.66                     & 33.89                    & 39.57                     & 33.59                    & \underline{37.80}                    & \underline{44.16}                     & 29.19                    & 33.87                    & 40.17                     & 29.86                     & 33.69                    & 39.69                    \\
LASA                 &                           & \underline{42.64}                    & \underline{45.29}                    & 54.26                     & \underline{35.17}                    & \underline{39.70}                    & 41.45                     & \underline{31.95}                     & 34.13                    & 35.14                     & \underline{35.66}                    & 33.79                    & 37.34                     & \underline{29.99}                    & 36.38                    & \underline{47.75}                     & \underline{29.87}                     & 34.40                    & \underline{48.27}                    \\
\name                  &                           & \textbf{44.68}           & \textbf{49.93}           & \textbf{57.40}            & \textbf{37.62}           & \textbf{41.25}           & \textbf{50.26}            & \textbf{33.69}            & \textbf{41.27}           & \textbf{51.15}            & \textbf{37.52}           & \textbf{42.77}           & \textbf{46.32}            & \textbf{34.95}           & \textbf{40.98}           & \textbf{48.37}            & \textbf{35.17}            & \textbf{39.38}           & \textbf{52.25}           \\ \hline
Krum                 & \multirow{9}{*}{CIFAR100} & 17.41                    & 22.64                    & 27.12                     & 17.24                    & 21.11                    & 25.68                     & \multicolumn{1}{c}{13.25} & 16.57                    & 21.26                     & 16.57                    & 18.82                    & 22.17                     & 15.84                    & 19.47                    & 24.24                     & 13.70                     & 17.13                    & 22.97                    \\
MKrum                &                           & 20.04                    & 26.55                    & 30.08                     & 16.71                    & 20.87                    & 24.54                     & 15.22                     & 17.07                    & 22.88                     & 13.51                    & 17.59                    & 23.55                     & 13.83                    & 19.06                    & 23.82                     & 15.43                     & 21.84                    & 25.19                    \\
Bulyan               &                           & 19.24                    & 21.55                    & 29.95                     & \underline{22.19}                    & 26.66                    & \underline{30.63}                     & 15.04                     & 17.52                    & 23.01                     & \underline{17.80}                    & 19.06                    & 23.82                     & 16.75                    & 22.54                    & 26.49                     & \underline{18.97}                     & 20.76                    & \underline{25.38}                    \\
Median               &                           & 19.51                    & 22.18                    & 30.21                     & 20.12                    & 22.13                    & 25.85                     & \multicolumn{1}{c}{15.65} & 18.02                    & 24.06                     & 17.34                    & 19.86                    & 24.36                     & 16.09                    & 22.40                    & 25.26                     & 14.55                     & 20.78                    & 25.12                    \\
Trmean               &                           & 18.36                    & 24.55                    & 32.58                     & 17.44                    & 19.03                    & 24.36                     & 15.23                     & 17.55                    & 23.34                     & 15.74                    & 17.36                    & 21.59                     & 14.24                    & 18.18                    & 23.95                     & 13.29                     & 18.17                    & 24.14                    \\
Dnc                  &                           & 21.62                    & 26.18                    & 30.61                     & 21.54                    & \underline{26.74}                    & 29.45                     & \multicolumn{1}{c}{\underline{19.56}} & \underline{23.21}                    & \underline{27.19}                     & 16.38                    & \underline{26.21}                    & \underline{30.19}                     & 15.25                    & \underline{25.03}                    & 27.87                     & 16.58                     & \underline{22.40}                    & 25.26                    \\
FLDetecter           &                           & 22.27                    & 26.00                    & 31.53                     & 20.18                    & 23.23                    & 27.80                     & 17.55                     & 20.94                    & 24.66                     & 17.16                    & 21.94                    & 26.28                     & 16.74                    & 21.11                    & 25.38                     & 14.35                     & 19.85                    & 24.25                    \\
LASA                 &                           & \underline{23.67}                    & \underline{30.64}                    & \underline{39.94}                     & 19.03                    & 22.41                    & 26.05                     & 17.51                     & 21.11                    & 24.97                     & 17.54                    & 20.32                    & 25.19                     & \underline{17.16}                    & 24.95                    & \underline{29.00}                     & 13.74                     & 22.17                    & 25.29                    \\
\name                  &                           & \textbf{25.62}           & \textbf{33.91}           & \textbf{40.59}            & \textbf{26.78}           & \textbf{28.41}           & \textbf{31.46}            & \textbf{25.76}            & \textbf{28.08}           & \textbf{31.86}            & \textbf{25.40}           & \textbf{27.84}           & \textbf{31.74}            & \textbf{23.91}           & \textbf{27.38}           & \textbf{29.13}            & \textbf{22.12}            & \textbf{26.94}           & \textbf{28.28}           \\ \hline
Krum                 & \multirow{9}{*}{FMNIST}   & 43.12                    & 53.66                    & 61.15                     & 44.64                    & 48.59                    & 59.32                     & 47.31                     & 52.64                    & 61.91                     & 43.10                    & 47.28                    & 57.03                     & 39.48                    & 45.51                    & 54.52                     & 39.97                     & 40.94                    & 48.54                    \\
MKrum                &                           & 44.41                    & 50.00                    & 61.66                     & 46.56                    & 51.32                    & 60.08                     & 48.58                     & 49.37                    & 57.44                     & 44.59                    & 46.74                    & 57.06                     & 40.80                    & 47.13                    & 55.70                     & 39.72                     & 44.61                    & 51.44                    \\
Bulyan               &                           & 42.91                    & 50.73                    & 59.45                     & \underline{48.39}                    & 52.28                    & 60.84                     & \textbf{51.68}                     & \underline{54.00}                    & 62.94                     & 45.63                    & 49.17                    & 58.49                     & \underline{42.06}                    & \underline{50.85}                    & \textbf{59.83}                     & 41.95                     & 50.37                    & 57.44                    \\
Median               &                           & 43.55                    & 49.76                    & 64.13                     & 46.74                    & 49.38                    & 57.49                     & 49.15                     & 53.72                    & \underline{63.09}                     & 44.79                    & 53.00                    & 60.86                     & 41.39                    & 48.68                    & 58.62                     & 41.88                     & 44.87                    & 52.26                    \\
Trmean               &                           & 43.62                    & 55.61                    & 62.53                     & 47.50                    & 52.81                    & 61.80                     & 48.78                     & 53.27                    & 61.80                     & 45.12                    & 51.59                    & 58.70                     & 40.32                    & 47.21                    & 53.49                     & 42.97                     & 49.32                    & 57.04                    \\
Dnc                  &                           & 47.97                    & 53.66                    & \textbf{64.34}                     & 47.57                    & \underline{54.46}                    & 62.87                     & 46.32                     & 49.55                    & 58.12                     & \textbf{46.52}                    & 53.38                    & 61.34                     & 40.21                    & 45.02                    & 53.38                     & \underline{44.23}                     & \underline{51.66}                    & \underline{60.38}                    \\
FLDetecter           &                           & 46.41                    & 58.78                    & 57.53                     & 47.77                    & 49.68                    & 57.96                     & 49.32                     & 52.45                    & 61.22                     & 44.40                    & 54.98                    & 63.37                     & 41.28                    & 45.49                    & 52.03                     & 41.83                     & 46.81                    & 56.07                    \\
LASA                 &                           & \underline{48.40}                    & \textbf{59.51}           & 59.66                     & 46.88                    & 53.72                    & \underline{63.09}                     & 49.67                     & 52.00                    & 60.86                     & 45.22                    & \underline{56.73}                    & \underline{65.93}                     & 41.44                    & 44.87                    & 52.26                     & 42.02                     & 50.38                    & 57.49                    \\
\name                  &                           & \textbf{49.54}           & \underline{59.37}                    & \underline{64.19}           & \textbf{49.07}           & \textbf{59.34}           & \textbf{67.35}            & \underline{50.86}            & \textbf{56.13}           & \textbf{65.70}            & \underline{46.43}           & \textbf{59.02}           & \textbf{66.92}            & \textbf{45.07}           & \textbf{51.32}           & \underline{59.35}            & \textbf{45.86}            & \textbf{53.49}           & \textbf{61.82}           \\ \hline
\end{tabular}
}
\label{table:robust_compare}
\end{table*}

In this section, we conduct experimental evaluations to measure the performance of the \name and its components.

\subsection{Experiment Settings}
\label{sec:Experimental_setup}

\noindent\textbf{Dataset:} We evaluate \name on three widely used datasets CIFAR-10~\cite{krizhevsky2009learning}, CIFAR-100~\cite{krizhevsky2009learning}, and FMNIST~\cite{xiao2017fashion}. To simulate realistic non-IID scenarios, each dataset is partitioned among clients using a Dirichlet distribution with parameter $\alpha$.
Before partitioning, we uniformly sample a fixed number of instances from each class to create a globally shared test set (global test set) for evaluating global accuracy across all classes. Each client then splits its local data into 80\% for training and 20\% for a local test set, used to assess performance on the client’s own distribution (local accuracy).

\noindent\textbf{Model:} Each client uses one of two heterogeneous model architectures: CNN-based model (\eg VGG) or RNN-based model (\eg vision-LSTM). This architectural heterogeneity reflects practical deployment scenarios. All client models are equipped with LoRA inserted into selected layers (\textit{feature-first} and the \textit{classifier}). The LoRA rank is set to 8 by default. Later, we will discuss in detail the choice of rank.

\noindent\textbf{Training and Poisoning:} Each communication round involves 10 clients, each performing 1 epoch of training with a batch size of 256. We run 200 rounds in total. Prior to FL, all clients undergo 10 epochs of local warm-up using their own data. Training uses standard SGD with dataset-specific learning rate tuning. The choice of $k{=}5$ follows the paper~\cite{eckart1936approximation}. 
To ensure fair comparison with baselines, 20\% of clients are selected in each round to perform poisoning attacks, which include both data and model poisoning, covering six types of attacks. All attacks begin from round 20.

\noindent\textbf{Baselines:} To comprehensively evaluate the performance of \name, we compare it with two groups baselines:
\begin{itemize}
\item Robust FL:
Verify whether \name can achieve comparable or superior robustness under attacks, including Krum~\cite{blanchard2017machine}, Multi-Krum~\cite{blanchard2017machine}, Bulyan~\cite{mhamdi2018hidden}, Median~\cite{yin2018byzantine}, Tr-mean~\cite{yin2018byzantine}, Dnc~\cite{shejwalkar2021manipulating}, FLDetector~\cite{zhang2022fldetector}, LASA~\cite{DBLP:conf/wacv/XuZH25}.

\item Heterogeneous and Efficient FL: Evaluate whether \name provides better performance under hyper-heterogeneity, including Fedrolex~\cite{alam2022fedrolex}, HeteroFL~\cite{diao2020heterofl}, MFL~\cite{DBLP:journals/network/YuL21}, Fjord~\cite{yu2023fjord}, FGGP~\cite{wan2024federated}, FedHello~\cite{chen2024fedhello}, FedALA~\cite{zhang2023fedala}.

\end{itemize}






For fair comparison, we use the same metrics as the baselines, namely average \textit{local accuracy} and \textit{global accuracy} cross-client~\cite{diao2020heterofl,shejwalkar2021manipulating}. All baseline implementations are faithfully reproduced based on their original papers, and we apply Dimensional Alignment to the clients' models for aggregation.

\subsection{Robustness Comparison}
\label{sec:Robustness}

To systematically evaluate the robustness of \name and baselines against poisoning attacks, we follow the experimental setup of the related work and adopt six attacks, including label-flip~\cite{bagdasaryan2020backdoor}, LIE~\cite{baruch2019little}, Min-Max~\cite{shejwalkar2021manipulating}, Min-Sum~\cite{shejwalkar2021manipulating}, Fang~\cite{fang2020local}, and AGR-tailored~\cite{shejwalkar2021manipulating}, and compare \name against a set of state-of-the-art robust FLs. Due to space constraints, we report only global accuracy in Table~\ref{table:robust_compare}, and the local-accuracy results exhibit the same trends. As shown in Table
~\ref{table:robust_compare}, across the 54 dataset/attack/$\alpha$ sub-columns reported in Table~\ref{table:robust_compare}, \name attains the highest global accuracy in 49 cases and places second in the remaining sub-columns, underscoring its strong robustness.

\begin{itemize}
\item On CIFAR-10, \name achieves the top result in every setting, yielding an average improvement of 2.73\% over the strongest baseline.
On CIFAR-100, \name also ranks first with a larger average gain of 3.29\%. Especially for Bulyan, The peak improvement 7.60\% appears at \emph{min-sum}, $\alpha{=}0.10$. These results indicate that \name’s robustness advantage.

\item On FMNIST, \name ranks first in 13/18 settings, and in the remaining 5 it places second.
This pattern likely arises because \name leverages the spectral structure of LoRA, which is more informative on texture-rich datasets (\eg CIFAR100), whereas FMNIST exhibits lower diversity and the model has higher top-$k$ energy, allowing conservative coordinate-wise trimming methods to edge out poisoning.
Nonetheless, the gaps are minimal (0.34\% on average), while \name’s average improvement remains 1.43\%.

\item Across all $\alpha$ settings, \name achieves the highest global accuracy in 49 of the 54 dataset–attack–$\alpha$ configurations, demonstrating strong robustness to non-IID data.
\end{itemize}

Overall, \name remains highly robust and competitive.

\begin{table}[!t]
\caption{Aggregation performance on three datasets}
\resizebox{\linewidth}{!}{
\renewcommand{\arraystretch}{1.15}
\scriptsize
\centering
\begin{tabular}{cllllll}
\hline
\multirow{2}{*}{} & \multicolumn{2}{c}{CIFAR10}                            & \multicolumn{2}{c}{CIFAR100}                           & \multicolumn{2}{c}{Fmnist}                             \\
                  & \multicolumn{1}{c}{local} & \multicolumn{1}{c}{global} & \multicolumn{1}{c}{local} & \multicolumn{1}{c}{global} & \multicolumn{1}{c}{local} & \multicolumn{1}{c}{global} \\ \hline
HeteroFL          & 70.49                     & 45.23                      & 50.28                     & 25.37                      & 87.82                     & 68.19                      \\
MFL               & 68.53                     & 46.48                      & 44.44                     & 19.63                      & 80.54                     & 58.47                      \\
Fjord             & 73.05                     & 48.22                      & 45.79                     & 20.91                      & 88.60                     & 69.48                      \\
FedRolex          & 76.38                     & 51.76                      & 48.72                     & 23.80                      & 83.94                     & 62.55                      \\
FedALA            & 69.91                     & 44.14                      & 47.54                     & 22.76                      & 86.57                     & 66.31                      \\
FGGP              & 70.79                     & 45.57                      & \underline{51.25}                     & \underline{26.43}                      & 85.52                     & 64.74                      \\
FedHello          & \textbf{78.06}            & \underline{55.95}                      & 50.41                     & 25.41                      & \underline{89.14}                     & \underline{70.42}                      \\
\name              & \underline{77.12}                     & \textbf{57.47}             & \textbf{51.97}            & \textbf{27.14}             & \textbf{89.47}            & \textbf{72.66}             \\ \hline
\end{tabular}
}
\label{tab:hetero} 
\vspace{-10pt}
\end{table}

\subsection{Aggregation Performance}

To evaluate the effectiveness of our Projection-Aware Aggregation under hyper-heterogeneity, we compare against representative heterogeneous FL baselines. When client architectures differ, for fairness, we pad structurally inconsistent layers to a common shape to ensure compatibility with baselines that assume homogeneous input to the aggregator. After each communication round, we evaluate every local model on both local and global test sets to measure their classification accuracy. Table~\ref{tab:hetero} reports the resulting average local and global accuracies. The datasets are partitioned across clients using a Dirichlet distribution with $\alpha=0.5$. We also tested other $\alpha$ and observed consistent rankings and trends (omitted for space). From the results, we observe: 

\begin{itemize}
\item Across all three datasets, our method achieves the best global accuracy. Compared with the strongest baseline, our method has achieved varying degrees of improvement on all three datasets. The average improvement on the three datasets is 1.82\%.

\item For the local metric, \name achieves the best results on CIFAR‑100 and FMNIST, and is only 0.94\% short of the best method on the dataset on CIFAR‑10. Notably, this small local gap on CIFAR‑10 comes with a substantial global gain (1.52\%), indicating that our approach is better at global generalization rather than relying on purely personalized improvements. This benefit is aligned with our design that aggregates updates on the selected feature‑first layer, which captures more universal features shared across clients.

\item For the Local–Global gap, our method narrows the gap on CIFAR‑10 (19.65\%) and FMNIST (16.81\%), and is on par with the best baseline on CIFAR‑100. These results indicate that the proposed projection‑aware selective aggregation leads to higher performance in the global accuracy, while maintaining competitive local performance under multi‑heterogeneity.

\end{itemize}

\subsection{Convergence}
\label{sec:convergence}

\begin{figure}[t]
  \centering
  \begin{subfigure}[b]{0.48\linewidth}
    \centering
    \includegraphics[width=\linewidth]{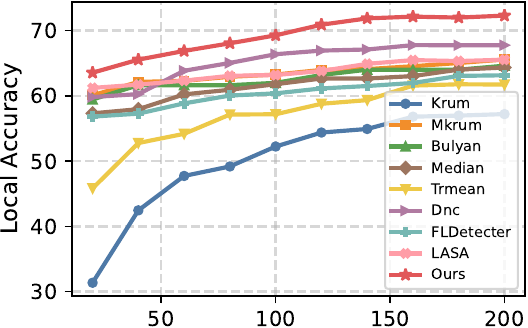}
    \caption*{}
    \phantomsubcaption\label{fig:rank}
  \end{subfigure}\hfill
  \begin{subfigure}[b]{0.48\linewidth}
    \centering
    \includegraphics[width=\linewidth]{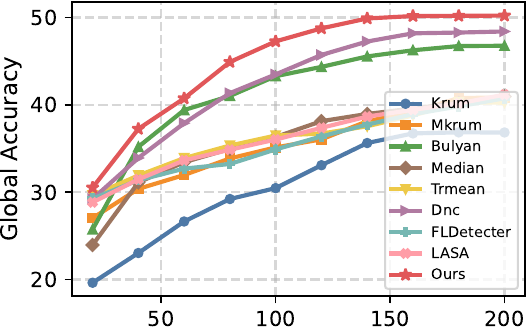}
    \caption*{}
    \phantomsubcaption\label{fig:lambda}
  \end{subfigure}
  \vspace{-15pt}
  \caption{Convergence analysis}
  \label{fig:two_parts}
  \vspace{-10pt}
\end{figure}

This section evaluates the convergence of our method compared to baseline models in heterogeneous federated learning settings.  Specifically, as in Fig. \ref{fig:two_parts}, we examine the number of communication rounds required to reach a target accuracy, and we plot the curves of “Communication Rounds vs. Global or Local Accuracy” to compare the convergence speed of different methods. From the Global Accuracy and Local Accuracy trajectories, we have observations:

\begin{itemize}
\item \textit{Faster convergence.} Our method quickly reaches a high-accuracy plateau in the early rounds, clearly outperforming the other robust aggregation baselines. Bulyan and LASA follow behind, while Krum/Mkrum, Median, and Trmean converge noticeably slower.

\item \textit{Stability under heterogeneity and attacks.} Once reaching the plateau, \name exhibits smaller oscillations, indicating stronger robustness and stability.  Other methods exhibit larger fluctuations and lower plateaus, underscoring their limited ability to resist attacks under multi‑heterogeneity.

\end{itemize}

\subsection{Communication Cost}

Communication cost is driven by two factors in synchronous FL: the number of effective rounds and the bytes transmitted per round, so we compare methods from two perspectives: \one the payload per round and \two the rounds needed to reach a target accuracy. As shown in \S\ref{sec:convergence}, our approach converges earlier and achieves higher accuracy, indicating that it requires fewer rounds at the same accuracy compared with the baselines. Thus, this section focuses on the per‑round upload payload (MB), consistent with the evaluation methodology used in the baseline.~\cite{diao2020heterofl,chen2024fedhello}.

\begin{figure}[t]
\centering
\includegraphics[width=0.75\columnwidth]{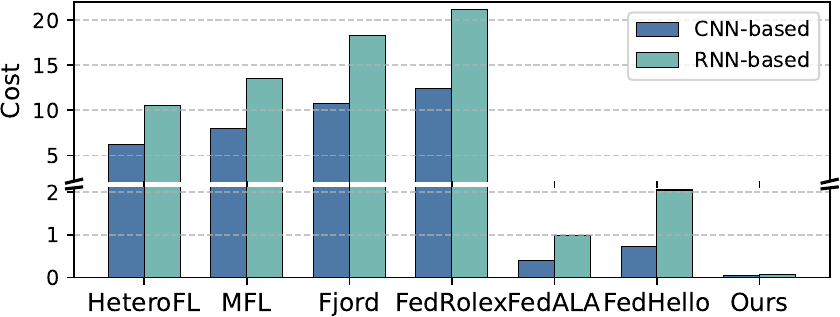} 
\caption{Communication cost}
\label{fig:communication_cost}
\vspace{-10pt}
\end{figure}

For the convenience of presentation and comparison of results, we present the costs of CNNs and RNNs, respectively. For FedALA, FedHeLLo, and \name, they operate under a lightweight regime where the total upload payload across all clients per round is in the sub-megabyte range (\ie less than 1MB). In contrast, other methods incur multi-megabyte per-client costs, typically ranging from 6–21MB per client per round, and we follow their original settings and report the per-type client per-round cost.
As shown in Fig.~\ref{fig:communication_cost}, our method incurs only 0.05MB/round for CNNs, achieving up to a 7.8× reduction in payload compared to the strongest lightweight baseline (FedALA), and up to a 420× reduction against high-overhead baselines like FedRolex, all without sacrificing performance (Table~\ref{tab:hetero}).
Similar trends are observed for RNNs, where our method achieves 12.5× to 271× lower payload across the evaluated baselines. Combined with faster convergence, the communication to reach the same accuracy is further reduced. This advantage becomes even more pronounced with larger models or stronger architecture heterogeneity.

\subsection{Ablation Study}
\label{sec:ablation}

\begin{figure}[t]
\centering
\includegraphics[width=0.8\columnwidth]{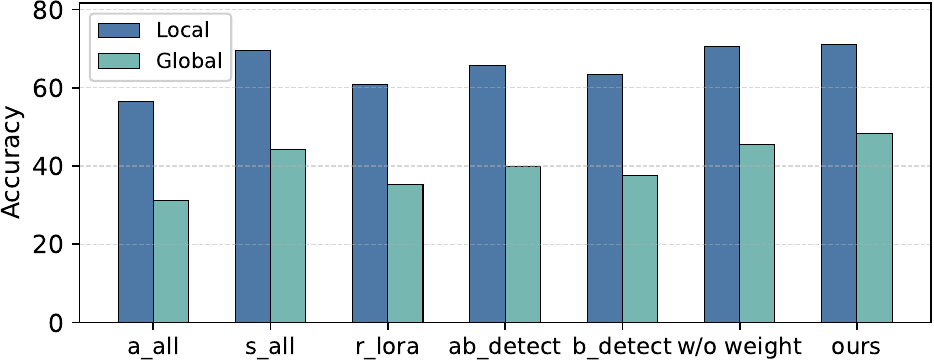} 
\vspace{-5pt}
\caption{Ablation study on six variants of \name}
\label{fig:ablation}
\vspace{-15pt}
\end{figure}

To verify the capabilities of each component of our method, we conducted an ablation study in three aspects: \one which to aggregate (all layers vs. selective layers vs. random layers), \two whether to decouple LoRA or not for detection (A/B/AB usage), and \three the usage of Projection-Guided Weighting. Table~\ref{tab:variant} shows the realization of the variants of \name. In addition to the change methods introduced in the table, all other settings are fixed. Fig.~\ref{fig:ablation} reports local and global accuracies of all variants under the Fang attack on CIFAR10, and we observe the same trends under other attacks. 

\textbf{Layer selection}: To verify our choice of aggregating the \textit{feature-first} and \textit{classifier}, we use different layer selection strategies in the aggregation process: \one full‑parameter aggregation on all layers (a\_all), \two LoRA aggregation on two random layers (r\_lora), and \three full‑parameter aggregation on our selected key layers (s\_all). For the method using full parameters, during the poisoning detection process, the full parameter update matrix is used for SVD decomposition to obtain its spectral correlation features for detection. We observe in Fig.~\ref{fig:ablation} that:

\begin{table}[t]
\caption{The variants of our method}
{\renewcommand{\arraystretch}{1.2}
\begin{tabular}{c|c}
\hline
name       & method                                                       \\ \hline
a\_all    & robust FL with all layer and all their parameters        \\
s\_all      & robust FL with selective layer and all their parameters \\
r\_lora     & robust FL with random two layers and their LORA          \\
b\_detect  & poisoning detection with LORA B of selective layer \\
ab\_detect & poisoning detection with LORA of selective layer    \\
w/o weight & aggregation without the projection-guided weight             \\ \hline
\end{tabular}
}
\label{tab:variant}
\vspace{-5pt}
\end{table}

\begin{itemize}

\item s\_all outperforms a\_all by 12.94\% in global accuracy and 13.25\% in local accuracy, indicating that aggregating only key layers (\ie, \textit{feature-first} and \textit{classifier}) significantly enhances robustness under adversarial settings. Such a decrease can be attributed to the fact that full-parameter aggregation leads to high-dimensional updates. On one hand, s\_all increases the attack surface, and on the other hand, such dispersed updates induced by the middle layer are harder to align with the principal direction of learning, making it difficult to consolidate global information, which ultimately degrades accuracy.

\item Compared to r\_lora, which randomly selects middle layers for aggregation, s\_all yields more stable performance because mid-level layers typically have less concentrated singular spectra and more drift-prone directions, leading to noisier signals for both detection and aggregation. In contrast, the first and last layers provide semantically meaningful, well-aligned features across clients. 

\item Our method still outperforms s\_all. This is because s\_all aggregates all parameters from the selected layers without any further filtering or selection, which increases the attack surface and introduces redundant or noisy directions into the aggregation process. Moreover, it directly performs SVD on the full parameters for poisoning detection. Based on our observations, instability can lead to false positives, thereby hindering the aggregation between normal clients. This reason will also be verified in the following chapters.


\end{itemize}

In summary, the results clearly demonstrate that selecting only stable and universal layers (\textit{feature-first} and \textit{classifier}) for aggregation significantly improves both robustness and accuracy under hyper-heterogeneity and attacks, validating our design choice in layer selection. Meanwhile, the addition of LoRA has also reduced the attack surface and enhanced the robustness of federated learning.

\textbf{LORA component for detection: }
To investigate the effectiveness of the components of LoRA in poisoning detection, we conduct an ablation study by using three strategies for detection: using only LoRA-A (\name), only LoRA-B (b\_detect), and both A and B (ab\_detect). All settings remain unchanged except for the input matrices used in HOPS.

Fig.~\ref{fig:ablation} shows Fang attack as an example, where we observe:
\begin{itemize}
\item Using LoRA-A (\name ) yields the best performance, achieving 48.37\% global accuracy and 70.98\% local accuracy, outperforming other variants.
\item LoRA-B alone performs worse than A-only, with global accuracy dropping to 37.60\% and local accuracy to 63.35\%. The reason is that B is more sensitive and unstable, and overreacts to benign fluctuations, leading to false positives and over-pruning of healthy clients, which degrades global learning.
\item Combining A and B (ab\_detect) is not better than using A alone. It achieves 39.77\% global accuracy and 65.71\% local accuracy, which is lower than A-only by 8.6\% and 5.27\%, respectively. Although combining A and B provides more information, the inherent instability of B contaminates the cleaner signal from A, leading to noisy or diluted detection scores.
\end{itemize}

These findings indicate that stability and consistency of the detection signal matter more than signal quantity, and using LoRA-A alone provides the best trade-off between detection robustness and accuracy.

\textbf{Projection-guided weighting:}
To examine the effectiveness of our proposed projection-aware weighting mechanism, we compare our complete method (\name ) with a variant that removes the weighting strategy (w/o weight). Without any guidance from projection, w/o weight aggregates selected LoRAs uniformly by Fedavg~\cite{mcmahan2017communication}, a classical aggregation method. As shown in Fig.~\ref{fig:ablation}:

\name  achieves 48.37\% global accuracy and 70.98\% local accuracy, while w/o weighted drops to 45.43\% global and 70.60\% local. Although the drop in local accuracy is small (only 0.38\%), the global accuracy drops significantly by 2.94\%, confirming that projection-aware weighting is critical for enhancing generalization than personalization. This is due to that, heterogeneous updates are averaged by w/o weight, even if they deviate from the principal optimization direction, which leads to unstable aggregation and diluted global learning. Our projection-guided weights filter noisy or misaligned components and amplify updates that contribute consistently along stable directions, better preserving the global learning signal, which is especially beneficial under attacks.

\subsection{Parameter Analysis}


We assess how the LoRA rank $r$ affects robust FL because $r$ is related to both model accuracy and communication (the communication payload grows roughly linearly with $r$), and it also impacts robustness: higher ranks enlarge the feasible update subspace, making adversarial perturbations easier to hide in high‑dimensional directions. We therefore we analyze our method's performance with different ranks, and in Fig. \ref{fig:rank}, we plot global accuracy versus rank under the LIE attack, and the cost curve belongs to RNNs. Other attacks exhibit the same pattern. From the Fig. \ref{fig:rank} we observe:

\begin{itemize}

\item Global accuracy exhibits a rise–saturation–drop pattern as the LoRA rank increases: from 
$r{=}4$ to $r{=}8$, it improves notably (2.42\%), and reaches the best at $r{=}16$, but the improvement is not obvious, only higher than $r=8$ by 0.35\%. Then declines when  $r\geq32$ (-1.44\% at $r{=}32$ and -2.08\% at $r{=}64$). This indicates that the intrinsic update subspace is low‑dimensional, and a moderate rank already captures the principal directions.

\item Too small $r$ (underfitting): capacity is insufficient to model the cross‑client principal directions, limiting global accuracy (e.g., $r{=}4$).
Moderate $r$ \ie $r{=}8/16$ balances expressiveness and noise suppression, yielding the best/near‑best accuracy.

\item A large rank $r$ spreads updates across many unnecessary directions, reducing aggregation consistency under heterogeneity. Moreover, it enlarges the feasible subspace for attacks, allowing poisoned biases to "hide" within heterogeneity-induced dynamics, thereby degrading global accuracy. For example, under a LIE attack, an adversary can exploit the increased degrees of freedom in the matrices to distribute small perturbations across more dimensions. This enables each coordinate’s deviation to remain within the defense’s “safety” thresholds, while collectively inducing a stronger overall bias.

\end{itemize}

In a nutshell, If communication and robustness are priorities, $r{=}8$ offers 50\% parameter/communication savings with only 0.35\% loss vs. the best. For highest accuracy, $r{=}16$ is a default. Thus, we choose rank=8.

\subsection{Discussion}



\begin{figure}[t]
  \centering
  \begin{subfigure}[b]{0.49\linewidth}
    \centering
    \includegraphics[width=\linewidth]{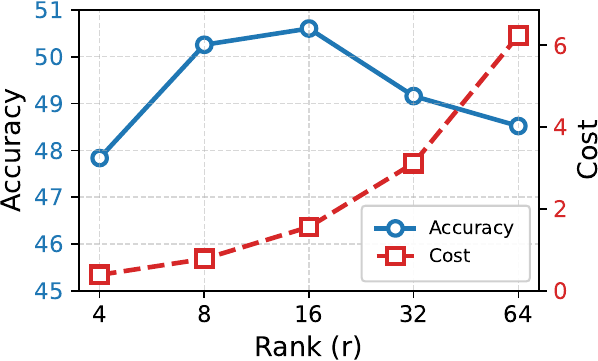}
    \caption*{}
    \phantomsubcaption\label{fig:rank}
  \end{subfigure}\hfill
  \begin{subfigure}[b]{0.49\linewidth}
    \centering
    \includegraphics[width=\linewidth]{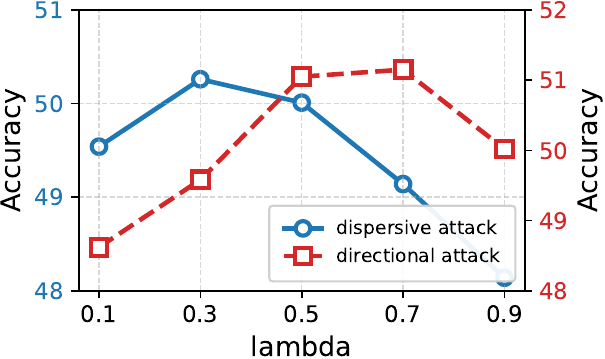}
    \caption*{}
    \phantomsubcaption\label{fig:lambda}
  \end{subfigure}
  \vspace{-15pt}
  \caption{Rank (left) and $\lambda$ (right) analysis.}
  \label{fig:rank_lambda}
  \vspace{-15pt}
\end{figure}

We sweep the HOPS coefficient $\lambda$ in Eq. (\ref{equa:HOPS}) under dispersive (\eg LIE/label-flip) and directional (\eg min-max/min-sum) attacks across clients and settings. Fig. (\ref{fig:lambda}) shows the global accuracy on CIFAR10-LIE-min-max as an example for brevity, and we observe the same qualitative trends across datasets and attacks. We can conclude that:

\begin{itemize}

\item Directional attacks achieve optimal accuracy on $\lambda=0.7$, as their poisoned updates are strongly aligned along a few principal directions, \ie $\mathcal{R}_k \uparrow$ and $\mathcal{H} \downarrow$. Hence, HOPS uses a larger $\lambda$ to emphasize the energy $\mathcal{R}_k$, making such low-rank, directional deviations easier to detect.

\item For dispersive attacks, perturbations are spread across many directions, yielding $\mathcal{R}_k \downarrow$ and $\mathcal{H} \uparrow$. In this case, a smaller $\lambda{=}0.3$ is preferred to upweight entropy and better capture dispersion.
\end{itemize}

Although accuracy varies with $\lambda$, \name remains consistently above baselines, indicating the gains come from the \name itself rather than tuning a single hyperparameter. We leave adaptive $\lambda$ to future work, promising options include narrowing the search with priors or online/meta-learning~\cite{DBLP:journals/corr/abs-1905-07435,DBLP:conf/gecco/GijsbersPRBV21} to adjust $\lambda$ per round to the prevailing attack pattern.

\section{Conclusion}

Federated learning is inherently vulnerable to poisoning attacks, a risk exacerbated by hyper-heterogeneity and high model dimensionality. To address this, we propose \name, a heterogeneity-oblivious robust FL framework centered on LoRA. Specifically, \name equips clients with LoRAs at two empirically stable layers and only shares LoRAs, effectively reducing the attack surface.
Leveraging an observed stability gap between LoRA-A and LoRA-B, the more stable LoRA-A is used to compute a heterogeneity-oblivious poisoning score for poisoning detection. For clients identified as benign, we perform projection-aware aggregation on LoRAs, reweighting updates based on principal-direction consistency to enhance both robustness and accuracy.
Empirical results demonstrate that \name consistently achieves state-of-the-art robustness and accuracy under diverse attack and heterogeneity conditions.

\bibliographystyle{abbrv}
\bibliography{bibfile}

\end{document}